\documentclass[letterpaper, 9 pt, conference]{ieeeconf}  

\IEEEoverridecommandlockouts                              

\makeatletter
\let\NAT@parse\undefined
\makeatother
\usepackage[hidelinks]{hyperref} 
\usepackage{xcolor}
\hypersetup{
colorlinks=false,
linkcolor=black,
citecolor=green,
}
\usepackage{cite}
\usepackage{graphicx}
\graphicspath{ {assets/} }
\usepackage{amsmath} 
\usepackage{amssymb}  
\usepackage{algorithm}
\usepackage{algorithmicx}
\usepackage{algpseudocode}
\usepackage{dsfont}
\usepackage{etoolbox}
\usepackage{amsbsy}                                
\usepackage{todonotes}
\usepackage{sidecap}

\errorcontextlines\maxdimen

\makeatletter
\newcommand*{\algrule}[1][\algorithmicindent]{\makebox[#1][l]{\hspace*{.5em}\vrule height .75\baselineskip depth .25\baselineskip}}%

\newcount\ALG@printindent@tempcnta
\def\ALG@printindent{%
    \ifnum \theALG@nested>0
        \ifx\ALG@text\ALG@x@notext
            \addvspace{-3pt}
        \else
            \unskip
            \ALG@printindent@tempcnta=1
            \loop
                \algrule[\csname ALG@ind@\the\ALG@printindent@tempcnta\endcsname]%
                \advance \ALG@printindent@tempcnta 1
            \ifnum \ALG@printindent@tempcnta<\numexpr\theALG@nested+1\relax
            \repeat
        \fi
    \fi
    }%
\usepackage{etoolbox}
\patchcmd{\ALG@doentity}{\noindent\hskip\ALG@tlm}{\ALG@printindent}{}{\errmessage{failed to patch}}
\makeatother

\title{\LARGE \bf
Uncertainty Averse Pushing with \\ Model Predictive Path Integral Control
}

\author{Ermano Arruda$^{*1}$, Michael J Mathew$^{*1}$, Marek Kopicki$^{1}$, Michael Mistry$^{2}$, Morteza Azad$^{1}$ and Jeremy L Wyatt$^{1}$
\thanks{*Joint first authors.}
\thanks{We gratefully acknowledge support of the Commonwealth scholarship by the British Council for Michael Mathew and a scholarship from the Brazilian National Council for Scientific and Technological Development (CNPq) for Ermano Arruda.}
\thanks{$^{1}$School of Computer Science,
        University of Birmingham, B15 2TT, UK
        {\tt\small (exa371,mjm522,msk,m.azad,jlw)@cs.bham.ac.uk}}%
\thanks{$^{2}$Edinburgh Centre for Robotics, University of Edinburgh,
        Edinburgh, EH8 9YL, UK
        {\tt\small mmistry@inf.ed.ac.uk}}%
}

\begin{document}

\maketitle
\thispagestyle{empty}
\pagestyle{empty}

\section{INTRODUCTION}
\begin{abstract}
Planning robust robot manipulation requires good forward models that enable robust plans to be found. This work shows how to achieve this using a forward model learned from robot data to plan push manipulations. We explore learning methods (Gaussian Process Regression, and an Ensemble of Mixture Density Networks) that give estimates of the uncertainty in their predictions. These learned models are utilised by a model predictive path integral (MPPI) controller to plan how to push the box to a goal location. The planner avoids regions of high predictive uncertainty in the forward model. This includes both inherent uncertainty in dynamics, and meta uncertainty due to limited data. Thus, pushing tasks are completed in a robust fashion with respect to estimated uncertainty in the forward model and without the need of differentiable cost functions. We demonstrate the method on a real robot, and show that learning can outperform physics simulation. Using simulation, we also show the ability to plan uncertainty averse paths. 
\end{abstract}



\section{INTRODUCTION}
Manipulating objects via non-prehensile actions, such as pushing, is a well-known problem in robotics \cite{dogar2011framework, cosgun2011push}. Planning these push-manipulations requires a forward model. There are many ways to express and acquire such a model, from analytic mechanics to machine learning, as well as hybrid techniques. There are several open problems, of which we address two. First, push planning typically does not take account of all the types of uncertainty in the predictions of the forward model. Second, when using purely learned models, push planning has only been demonstrated for single pushes, not for complex push sequences. In this work we present a combined solution to these problems. 



Uncertainty in prediction comes from two sources. First, it can arise from small variations in physical properties, such as shape and friction, that are hard to measure, but which significantly alter action effects. Second, in a learned forward model it can arise from a paucity of data. In this paper, we use two different learning methods to explicitly predict a distribution over push outcomes, including both types of uncertainty.

When planning an action sequence the robot can take account of regions of high uncertainty. This is important because actions in uncertain regions of the state space can lead to unrecoverable states. We model these as incurring a cost that rises with the uncertainty predicted by the forward model. But, what push planner can we use? The choice is complicated by the fact that the overall cost function is typically not a differentiable function of the actions. In this case, a path integral formulation \cite{kappen2005path,Theodorou2010,williams2015model} works well. We also utilise re-planning each step to account for model inaccuracies. Thus, our planner is a model predictive path integral (MPPI) controller that performs uncertainty averse pushing.

\begin{figure}[t]
\centering
\includegraphics[width=0.45\textwidth]{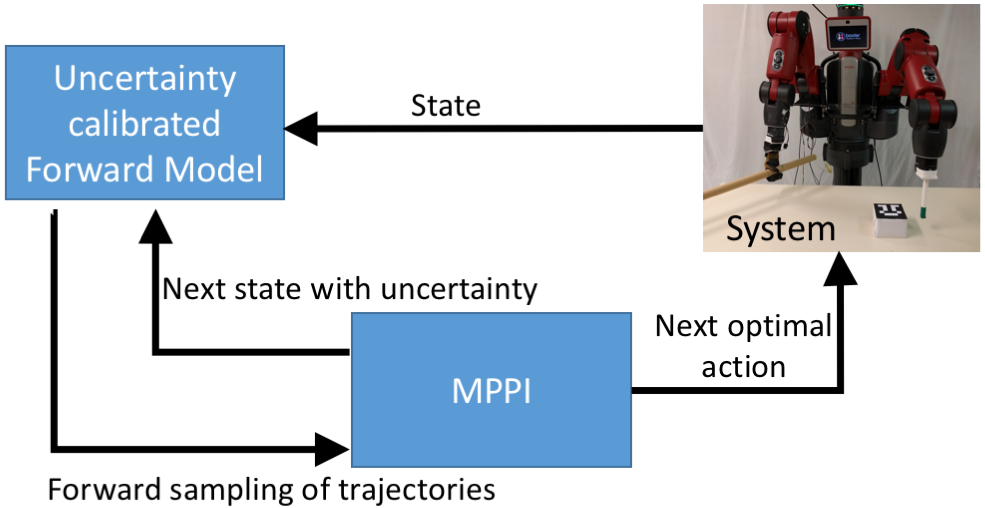}
\caption{\label{fig:path_A} A high-level diagram of the approach. The forward model is acquired from data, and gives uncertainty in its predictions. The system is the robot plus box and environment.}
\end{figure}  

The technical contributions of this work are: (i) applying ensembles of MDNs and Gaussian Processes to learn uncertainty aware forward models of push manipulation; (ii) using a learnt forward model with model predictive path integral (MPPI) control to push an object to a given goal pose with a many step plan; and (iii) two algorithms for uncertainty averse push planning.

The paper is organised as follows. Section~\ref{relatedWork} reviews existing work on push learning and planning. Section ~\ref{preliminaries} gives a brief background on the elements of our approach. Section~\ref{approach} explains the problem formulation and algorithms. Section~\ref{experiment} details the experimental study. Finally, Section~\ref{discussion} is a discussion.

\section{RELATED WORK}
\label{relatedWork}
\subsection{Learning Models for Pushing}

There are many approaches to modelling push effects based on classical analytic mechanics \cite{mason_manipulator_1982,lynch_mechanics_1992,peshkin_motion_1988,mason_mechanics_2001}. These require modelling and knowledge of parameters such as friction, mass, inertia, and centre of gravity, of the manipulated object, manipulator, and other objects in contact. Accurate identification of these parameters is hard. Even if this is solved, approximations used in rigid body engines can render poor predictions. An alternative is to learn a model from data \cite{Finn16,Agrawal16,PintoG16,Kopicki2016}, or to use a hybrid approach \cite{zhou2017fast,belter2014iros}. Learning methods divide into data-intensive and data-efficient methods.

Data-intensive methods, such as deep-learning, adopt self-supervision for data collection, allowing the creation of large datasets. In \cite{Agrawal16}, for example, a \textit{siamese} architecture learns a forward and an inverse model for pushing. The forward model is used as a regulariser for the inverse model. Limitations are the use of a discrete action space and the lack of a representation of predictive uncertainty. Finn et al \cite{Finn16} used an auto-encoder based forward model that is used in a model predictive control schema to find push actions based on image input. However, this model also lacks knowledge of the predictive uncertainty and has only been shown to achieve single step push manipulations. Pinto et al. \cite{PintoG16}, used push models to improve grasp performance through pushing. Again, predictive uncertainty in the learnt models was not explored. 

There are also data-efficient approaches to learning push effects \cite{Kopicki2016,kopicki2011learning,bauza17}. These models typically use hand-crafted features, and have not yet been used for push planning. They do, however, represent uncertainty in the outcome, including, in \cite{bauza17}, the ability to predict meta-uncertainty due to a lack of data.

\subsection{Estimating Predictive Uncertainty}


The ability to predict both uncertainty in dynamics and meta-uncertainty in this dynamics model is useful for robot planning \cite{deisenroth2011pilco}. Policy search method PILCO \cite{deisenroth2011pilco} utilises Gaussian Processes (GPs) \cite{williams2015model}. GPs, however, scale poorly with the amount of training data. Representations such as Gaussian Mixture Regression scale better and can estimate dynamic uncertainty, but not the meta uncertainty \cite{da2012learning}. A neural network approach to representing uncertainty in dynamics is to learn the parameters of a mixture density. This is termed an mixture density network (MDN). This can be extended to add meta-uncertainty due to a lack of data in various ways. One way is to use dropout \cite{gal2016dropout}. Another way is to use an ensemble of MDNs and adversarial training \cite{lakshminarayanan2016simple}. We use this latter approach.

\subsection{Path Integral Applications for Control}

Stochastic optimal control (SOC) deals with both uncertainty in the action and sensor models, and the resultant state uncertainty.  The sequence of control commands is found by minimising an integral of individual step costs (called the running cost) along a given trajectory.  
The SOC problem is defined by a \textit{Hamilton-Bellman-Jacobi}  (HJB) partial differential equation (PDE) corresponding to the system to be controlled. This can be solved numerically backwards in time, given the system's initial and target configurations \cite{kappen2005path}. This is straightforward for linear systems with quadratic costs\cite{li2004iterative}, but non-trivial for non-linear systems.

However, by using the Feynman-Kac theorem, a non-linear HJB can be converted into a linear PDE, which can be solved via forward sampling of trajectories \cite{Theodorou2010,kappen2005path}. This formulation can cope with arbitrary state costs that need not be differentiable, and is applicable to a wide range of non-linear systems. Recently, researchers have explored its benefits for robot control \cite{ChebotarKYLSL16,williams2015model,Williams2016}. We make use of the path integral framework. Specifically, we apply the model predictive path integral control algorithm proposed by \cite{williams2015model}.

\section{PRELIMINARIES}
\label{preliminaries}
Classical optimal control deals with finding a set of control actions that solves the problem (typically deterministic systems) and is optimal with respect to a cost function. The general framework is to design an agent as an automaton that seeks to minimise a cost function for a fixed or varying time horizon \cite{kappen2011optimal}. There are typically two methods to solve an optimal control problem. They are the HJB formulation (finding a solution using dynamic programming) and the second is by using \textit{Pontryagin's Minimum Principle} (PMP) (finding a solution to the ordinary differential equations formed) \cite{kirk2012optimal}. These formulations are typically interested in finding a globally optimal solution.
%
%
%
%

In this paper we define a trajectory as being a sequence of states $\mathbf{x}_t$, actions $\mathbf{u}_t$ and associated uncertainty $\mathbf{\hat \sigma}_{t,n}$ at time step $t$. Thus, let $\mathbf{s}_{t,n}$ be a convenient tuple, such that $\mathbf{s}_{t,n} = (\mathbf{x}_{t,n},\mathbf{u}_{t,n},\mathbf{\hat \sigma}_{t,n})$. Then, using $n$ to index a trajectory and $i$ to index a discrete timestep, the optimal control problem is cast by defining a cost function for a trajectory $n$ starting from timestep $i$ until $T$, i.e. $\tau_{i,n} = [\mathbf{s}_{i,n}, \mathbf{s}_{i+1,n},  \mathbf{s}_{i+2,n}, ..., \mathbf{s}_{T,n}]$:
\begin{equation}
S_i = S(\tau_{i,n})  = \phi_T(\mathbf{x}_T) + \sum_{t=i}^{T-1} r_t(\mathbf{x}_t,\mathbf{u}_t, \mathbf{\hat \sigma}_t)
\end{equation}
where, $r_t$ is the immediate cost function and $\phi_T$ is the final cost function. The aim is to find a policy that minimises the above cost function. The value function for a state is defined as the minimum cumulative cost the agent can obtain from a state, if it proceeds optimally from that state to the goal. The value function for a state $\mathbf{x}_i$ can be defined as:
\begin{equation}
\label{eq:value_func}
V(\mathbf{x}_i) = \min_{\mathbf{u}_{i:T}} E[S_i]
\end{equation}
At any state $\mathbf{x}_i$, the aim is to find a set of control commands or actions $\mathbf{u}=(\mathbf{u}_i, ... \mathbf{u}_T)$ that would minimise the expected cumulative cost from that state. Note that in Eq \ref{eq:value_func} the expectation is taken over all possible paths (i.e. trajectories) and thus the index $n$ is dropped for convenience.

\section{APPROACH}
\label{approach}
We now describe the proposed uncertainty averse push planner, which comprises two parts. First, an uncertainty calibrated forward model is learnt with data collected from a variety of pushes. Second, using the learnt model, we formalise our planning problem as Model Predictive Path Integral control (MPPI) and detail our planning algorithm. Later we describe another path integral based approach to find a low cost trajectory that can be followed by using the MPPI controller.

\subsection{Forward Models with Predictive Uncertainty}

There are various ways to capture uncertainty in predictions. Gaussian Processes, for example, provide an effective and theoretically clean tool to make uncertainty aware predictions \cite{rasmussen2006gaussian}. The main problem with Gaussian processes (GPs) is their inability to scale to high dimensional spaces or large data sets.  We therefore also utilise an ensemble of mixture density networks (E-MDN).

Arbitrary densities can also be learnt via gradient descent with Mixture Density Networks (MDNs) \cite{Bishop94}. In such models, if we choose the mixture components to be Gaussian, the network outputs the parameters of a Gaussian mixture model conditioned on a suitable choice of input vector $\mathbf{h} \in \mathcal{R}^n$, thus modelling arbitrary multi-modal densities as defined in equation \ref{eq:mixture}.

\begin{equation}
\label{eq:mixture}
p_{\theta}(\mathbf{x}_{t+1} | \mathbf{h};{\theta}) = \sum_{k=1}^{K} \pi_k(\mathbf{h};{\theta}) \mathcal{N}(\mathbf{x}_{t+1}|\mathbf{\mu_k(h;{\theta})},\sigma_k^2(\mathbf{h;{\theta}})),
\end{equation}  

where $\pi_k(\mathbf{h};{\theta})$, $\mu_k(\mathbf{h};{\theta})$ and $\sigma_k(\mathbf{h;{\theta}})$ are the network outputs which form the parameters of the mixture. In order to make sure the network outputs valid parameters for the mixture, Bishop \cite{Bishop94} suggests using a softmax layer to represent $\pi_k$, such that $\sum_{k=1}^K \pi_k(\mathbf{h};{\theta}) = 1$, whereas an exponential layer is able to guarantee that $\sigma_k$ is positive definite, and finally $\mu_k$ can be a linear combination of hyperbolic tangent activation functions. The reader is encouraged to refer to Bishop \cite{Bishop94} and Graves \cite{Graves13} for further details on practical considerations in implementing MDNs.

Concretely, for learning a forward model, we define $\mathbf{h} = [\mathbf{x}_{t},\mathbf{u}_t]$, where $\mathbf{x}_{t}$ is the state at time step $t$, (position and orientation of the box on the plane) $\mathbf{x}_{t} = [x_t,y_t,\theta_t]$,  and $\mathbf{u}_t$ is the action taken at that time, encoded as a direction vector [$px_t, py_t$] and a single real value $a_t$, normalised between zero and one,  indicating the contact location on the box edge, i.e. $\mathbf{u}_t = [px_t, py_t, a_t]$ (all quantities are given in the object frame). Furthermore, as has been demonstrated by \cite{lakshminarayanan2016simple}, one can form an ensemble of such models so as to estimate the uncertainty of the forward model's predictive distribution. If an ensemble is composed of $M$ members, the final model can be written as:

\begin{equation}
\label{eq:ensemble}
p(\mathbf{x}_{t+1}|\mathbf{x}_t,\mathbf{u}_t) = \frac{1}{M} \sum_{m=1}^{M} p_{{\theta}_m}(\mathbf{x}_{t+1}|\mathbf{x}_t,\mathbf{u}_t;{\theta}_m)
\end{equation}

The statistics of interest that we compute from the ensemble are the mean prediction and the variance, which, when trained accordingly, can reflect the predictive uncertainty of the model \cite{lakshminarayanan2016simple}, and are given by:

\begin{equation}
\label{eq:ensemble_mean}
\hat{\mu} = \frac{1}{M} \sum_{m=1}^{M} \mu_{\theta_m}
\end{equation}

\begin{equation}
\label{eq:ensemble_variance}
\hat{\sigma}^2 = \frac{1}{M} \sum_{m=1}^{M} (\sigma^2_{\theta_m} + \mu^2_{\theta_m}) - \hat{\mu}^2
\end{equation}

Thus, the predictive uncertainty we refer throughout this paper represents both inherent uncertainty in dynamics, and meta uncertainty due to lack of data, as given by \ref{eq:ensemble_variance}.


\subsection{Uncertainty Averse Model Predictive Path Integral Control}

Once we have a forward model, we need a way to use this model to find the right sequence of push commands to move the object to the goal. When solving a task, it is easier to exploit the already known part of the state-action space rather than exploring new parts. This suggests moving through more certain regions of the state-action space. We use the path integral based approach to model predictive control in \cite{williams2015model}.

The path integral formulation for stochastic optimal control permits one to find policy updates by calculating expectations over trajectory roll outs \cite{Theodorou2010}. It provides an alternative to directly solving the non-linear HJB equation via backward integration, and allows one to find the command updates that minimise the cost-to-go in Eq \ref{eq:cost_to_go} as a weighted average over $N$ forward sampled trajectories.

\begin{equation}
\label{eq:cost_to_go}
S(\tau_{i,n}) =  \phi(\mathbf{x}_T) + \sum_{t=i}^{T-1} r(\mathbf{x}_t,\mathbf{u}_t,\hat{\sigma}_t),
\end{equation}
Given the cost-to-go in Eq \ref{eq:cost_to_go}, one wants to find the optimal action sequence as $\overset{*}{\mathbf{u}} = \arg \min_{\mathbf{u}} E[S_i]$. Note that in this work the cost is also a function of the predictive uncertainty $\hat{\sigma}_t$, in addition to the state $\mathbf{x}_t \in \mathcal{R}^n$ and controls $\mathbf{u}_t \in \mathcal{R}^m$. The importance of each $n^{th}$ sample is given by a weight, defined as the exponential of the cost-to-go $S(\tau_{i,n})$, which is given by:

\begin{equation}
\label{eq:pi_weight}
w_{i,n} = \frac{\exp(-\frac{1}{\lambda} S(\tau_{i,n}))}{\sum_{l=1}^{N} exp(-\frac{1}{\lambda} S(\tau_{i,l}))}
\end{equation}

where $\lambda$ can be seen as the temperature parameter for the softmax distribution in Eq \ref{eq:pi_weight} and affects the control update given by Eq \ref{eq:pi_delta_u}. 

\begin{equation}
\label{eq:pi_delta_u}
\Delta \mathbf{u}_i = \sum_{n=0}^{N} w_{i,n} \delta \mathbf{u}_{i,n}(\eta)
\end{equation}

Thus, the control command updates are calculated as an expectation, or weighted average, over sampled control disturbances $\delta \mathbf{u}_{i,n}$ with weights equal to $w_{i,n}$. Here, control disturbances are in a similar manner to that  defined in \cite{williams2015model}. However, we introduce an exploration decay parameter $\eta$, i.e.

\begin{equation}
\label{eq:del_u}
\delta \mathbf{u}_{i,n}(\eta) = \eta \frac{1}{\sqrt{\rho}} \frac{\epsilon}{\sqrt{\Delta t}},
\end{equation}

where $\Delta t$ is the time step magnitude, with $\epsilon \sim \mathcal{N(\mathbf{0},\mathbf{I})}$, which has same dimensionality as the control actions $\mathbf{u_t} \in \mathcal{R}^m$. The parameter $\rho$ can be seen as a constant responsible for controlling the magnitude or level of exploration of the sampled disturbances $\delta \mathbf{u}_{i,n}$, whereas a suitable exploration decay schedule for $\eta$ helps to ensure local convergence even when the number $N$ of sampled trajectories is small. In all experiments presented in this paper $\eta$ is always initialised as $\eta=1.0$ and geometrically decays over a chosen number of decay steps $L$ as detailed in Algorithm \ref{alg:algorithm_1}.

By using a model predictive approach we are able to incorporate feedback into the system. At each state a look-ahead window of $T$ time steps is used, starting at time step $i$. Then the first control command of the $T$ steps is executed on the robot. After this first push, the new state is fed back into the optimiser and the process is repeated till task convergence. 

The immediate cost function used in our formulation is

\begin{equation}
 r(\mathbf{x}_i,\mathbf{u}_i,\hat{\sigma}_i) = \gamma * \hat{\sigma}_i  + (\mathbf{x}_i - \mathbf{x}_{goal})^T Q (\mathbf{x}_i - \mathbf{x}_{goal}) + \mathbf{u}_i^T R \mathbf{u}_i,
 \label{eq:mppi_cost}
\end{equation}

 and the final cost is given by
 
\begin{equation}
 \phi(\mathbf{x}_T) = (\mathbf{x}_T - \mathbf{x}_{goal})^T Q (\mathbf{x}_T - \mathbf{x}_{goal})
 \label{eq:mppi_final_cost}
\end{equation}

By adding the $\hat{\sigma}$ term in equation \ref{eq:mppi_cost}, the samples that pass through an uncertain region are penalized more and hence would contribute less to the control update in equation \ref{eq:pi_delta_u}. Throughout the remainder of the paper, the state of the object to be pushed is defined by its position and orientation on the plane under the quasi-static assumption, subject to only planar motion, i.e. $\mathbf{x}_i = [x_i,y_i,\theta_i]$.


\begin{algorithm}[h]
\caption{The modified MPPI algorithm with exploration decay, a modification of the the original algorithm proposed by \cite{williams2015model} \label{alg:algorithm_1} that uses an uncertainty calibrated forward model (Eq \ref{eq:ensemble_mean} and \ref{eq:ensemble_variance})}
\begin{algorithmic}
\State \textbf{Given}: 
\State N: Number of samples;
\State T: Number of timesteps;
\State L: Number of decay steps;
\State $(\mathbf{u}_1,\mathbf{u}_2, ..., \mathbf{u}_{T})$: Initial action sequence;
\State $\eta_{init}$: Initial $\eta$, for exploration decay, Eq \ref{eq:del_u};
\State ${\Delta} t, \mathbf{x}_{init}, \hat{\mu}(\cdot,\cdot), \hat{\sigma}(\cdot,\cdot)$: System sampling dynamics; 
\State $\phi, r, \mathbf{R}, \mathbf{Q}, \lambda, \gamma$: Cost parameters;
\State $\mathbf{u}_{init}$: Value to initialise new controls to;
\State
\While{\textit{task not completed}}
\State $\eta = \eta_{init}$
\For{$l = 0$ \textbf{to} $L$} 
\For{$n = 1$ \textbf{to} $N$} 
  \State $\mathbf{x_0} = \mathbf{x}_{init}$
  \State $\hat{\sigma_{0}} = \hat{\sigma}(\mathbf{x}_0,\mathbf{0})$ (given by Eq \ref{eq:ensemble_variance})
  \For{$i = 1$ \textbf{to} $T - 1$} 
  \State $\mathbf{x}_{i+1} = \hat{\mu}(\mathbf{x}_i,\mathbf{u}_i + \delta \mathbf{u}_{i,n}(\eta))$ (given by Eq \ref{eq:ensemble_mean})
  \State $\hat{\sigma_{i+1}} = \hat{\sigma}(\mathbf{x}_i,\mathbf{u}_i + \delta \mathbf{u}_{i,n}(\eta))$ (given by Eq \ref{eq:ensemble_variance})
  \State $S(\tau_{i+1,n}) = S(\tau_{i,n}) + r(\mathbf{x_i},\mathbf{u_i},\hat{\sigma_i})$
  \EndFor
  \State $S(\tau_{T,n}) = \phi(\mathbf{x_T})$
\EndFor
\For{$i = 1$ \textbf{to} $T$} 
\State $\mathbf{u}_i = \mathbf{u}_i + \Delta \mathbf{u}_i$ (with $\Delta \mathbf{u}_i$ given by Eq \ref{eq:pi_delta_u})
\EndFor
\State $\eta = 0.99 \eta$ (exploration decay)
\EndFor
\State send\_control\_command($\mathbf{u}_0$)

\For{$i = 1$ \textbf{to} $T - 1$} 
\State $\mathbf{u}_i = \mathbf{u}_{i+1}$
\EndFor
\State $\mathbf{u}_{T-1} = \mathbf{u}_{init}$

\State Update current state
\State Check task completion

\EndWhile
\end{algorithmic}
\end{algorithm}

\subsection{An Alternative Approach to Uncertainty Averse Planning}

The performance of MPPI for uncertainty averse planning depends on the cost function defined. The challenge of the cost function defined in equation \ref{eq:cost_to_go} is to optimise the trade off between $\mathbf{Q}$ and $\gamma$. 

There is a different approach that involves decoupling goal finding and uncertainty reduction. First, a simple path is defined from start to goal. Then the learnt forward model can be used to find a low uncertainty variation of this path. Finally, this can be given to the model predictive controller to follow. The decoupling of the uncertainty cost and the final goal cost gives a significant improvement in performance. We propose a path integral based approach to find a low cost trajectory in the state-action space that can then be followed by the MPPI. The formulation derives inspiration from the STOMP planner \cite{kalakrishnan2011stomp}, though our formulation uses a different cost function and improvement equations. Apart from finding a low cost trajectory, kinematic constraints are imposed on the optimisation problem to find paths within the workspace of the system. 

The problem of finding a low uncertainty trajectory can be formulated as:
\begin{eqnarray}
\label{eq:sigma_cost}
\begin{aligned}
\text{Minimise: }& S(\tau_{i,n}) = \alpha \sum_{i=0}^{T}  \hat{\sigma_i} \\
\text{Subject to state constraints: }& \mathbf{x}_{min} \leq \mathbf{x} \leq \mathbf{x}_{max}
\end{aligned}
\end{eqnarray}

where, for a state $\mathbf{x}$ at time $i$ the uncertainty predicted is $\sigma_i$ and $\alpha$ is a positive constant. The uncertainty for each point in the heat-map is estimated via Monte Carlo sampling. Thus, by iterating over a subset of states (e.g. states on a grid) and random actions for each state, the average uncertainty is estimated using the learnt forward model with Eq \ref{eq:ensemble_variance} for each state. With this uncertainty heat-map, Algorithm \ref{alg:algorithm_2} starts with a initial trajectory leading from start to goal. This could be simple linear interpolation from start to goal. If the total number of states in this interpolation is $T$, then the optimisation is performed for states from $1$ to $T-1$, thus ensuring the path continues to reach the goal from the start. The initial trajectory is improved iteratively by usinq Eq  \ref{eq:pi_weight} for the cost defined in Eq \ref{eq:sigma_cost}. The iterative update is defined by:

\begin{equation}
\label{eq:pi_delta_x}
\Delta \mathbf{x}_i = \sum_{n=0}^{N} w_{i,n} \delta \mathbf{x}_{i,n}
\end{equation}


\begin{algorithm}
\caption{Algorithm 2 optimises a trajectory with respect to model predictive uncertainty. The output of the algorithm is a trajectory with low uncertainty cost which can be tracked by a push controller, in this work we use the Model Predictive Path Integral Controller for following the pushing trajectory.}
\label{alg:algorithm_2}
\begin{algorithmic}
\State \textbf{Given}: 
\State N: Number of samples;
\State T: Number of time steps;
\State $f(\mathbf{x},\mathbf{u})$: Learnt forward dynamics;
\State $\epsilon$: Threshold of cost change;
\State $\tau$: $(\mathbf{x}_0,\mathbf{x}_2, ..., \mathbf{x}_{T-1})$;
\State $\alpha$: positive constant multiplier to cost;
\While{$S \geq \epsilon$}
\For{$n = 0$ \textbf{to} $N$} 
\For{$i = 1$ \textbf{to} $T-1$}
\State $\delta \mathbf{x}_{i,n} \sim \mathcal{N}(\mathbf{0},\mathbf{I})$
\State $\mathbf{x}_{i,n}$ = $\mathbf{x}_i + \delta \mathbf{x}_{i,n}$
\State Enforce\_state\_constraints($\mathbf{x}_{n,i}$)
\State $S(\tau_{i+1,n}) = S(\tau_{i,n}) + \frac{\alpha}{T} \hat{\sigma_i}$
\EndFor
\EndFor
\For{$i = 1$ \textbf{to} $T - 1$} 
\State $\mathbf{x}_i = \mathbf{x}_i + \Delta \mathbf{x}_i$ (with $\Delta \mathbf{x}_i$ given by Eq \ref{eq:pi_delta_x})
\State Enforce\_state\_constraints($x_i$)
\EndFor
\EndWhile
\end{algorithmic}
\end{algorithm}

\section{EXPERIMENTS}
\label{experiment}
Experiments 1 and 2 described below validate the fundamental ability to push objects to a goal location using E-MDN as the learnt forward model on a real robot (Baxter), without considering uncertainty in the cost function. Experiments 3 and 4 demonstrate in simulation that the same approach is able to find uncertainty averse trajectories for pushing an object to a given goal. For the experiments in the real robot, we found that uncertainty averse pushes achieved in simulation were sometimes hard to reproduce on the Baxter platform due to kinematic limitations of the robot, i.e. the inverse kinematic solver sometimes failed to find solutions for planned pushes. Thus, general pushing is shown on the real robot, whereas uncertainty averse pushing (which requires using a much larger workspace) is demonstrated in simulation only.  We now proceed to describe the experiments in the real robot. Finally, we will describe the simulation experiments that demonstrate uncertainty averse pushing.\footnote{See video summarising the proposed approach and experiments: \url{https://youtu.be/LjYruxwxkPM}}

\subsection{Experiment 1:} 
\begin{figure}[t]
\centering
\includegraphics[width=1.\columnwidth]{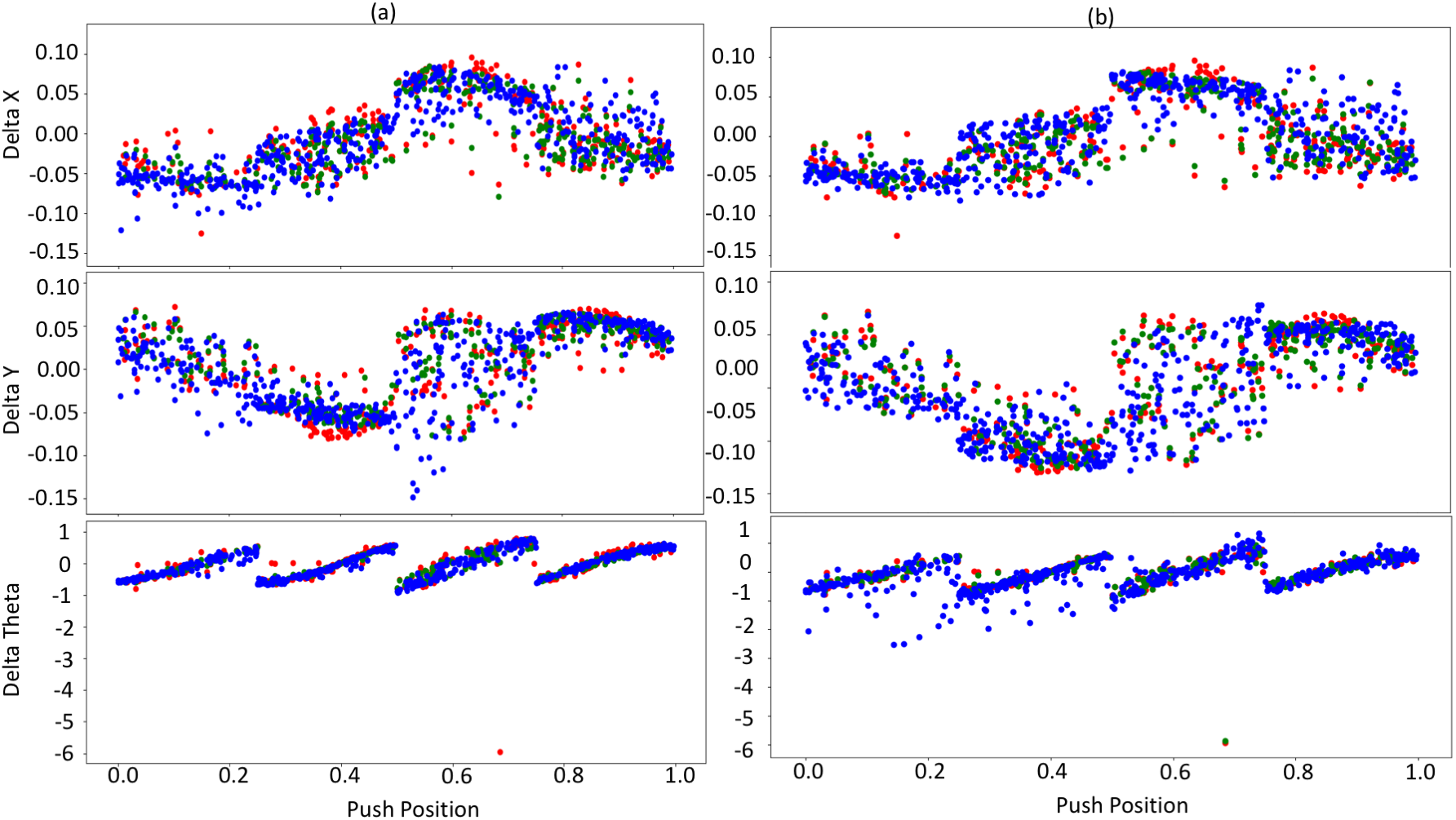}
\caption{\label{fig:unc-comp}(a) Predictions of the E-MDN.  (b) Predictions of the GP. The top row shows predicted X displacement. The middle shows Y displacement. Theta shows the change in orientation. The red dots illustrate the true delta, the green dots show the prediction on the training set. The blue dots show the prediction on a test set generated from a distribution having same mean and covariance as the training set.}
\end{figure}

In this experiment we trained the E-MDN model on a set of 326 pushes, gathered from the Baxter robot \footnote{see data collection setup, in which the robot applies random pushes to the object and periodically restarts the box to an initial location: \url{https://youtu.be/pRDvkDkCSTQ}}. The E-MDN utilised had $M=10$ members in the ensemble, each member being an MDN with 3 hidden layers with 20 neurons per layer. The number of mixtures in each MDN was chosen to be $K=1$ for simplicity. The model was trained for 3000 epochs with stochastic gradient descent using the Adam \cite{KingmaB14} optimiser. We utilised batch size 5 and the learning rate was 0.001. Following the training protocol described by \cite{lakshminarayanan2016simple}, we used 0.005 as the adversarial coefficient for generating adversarial examples for training. Figure~\ref{fig:unc-comp}(a) and (b) shows the predictions given by the trained E-MDN model and the GP model respectively. 

\begin{figure}[!h]
\centering
\includegraphics[width=1.0\columnwidth]{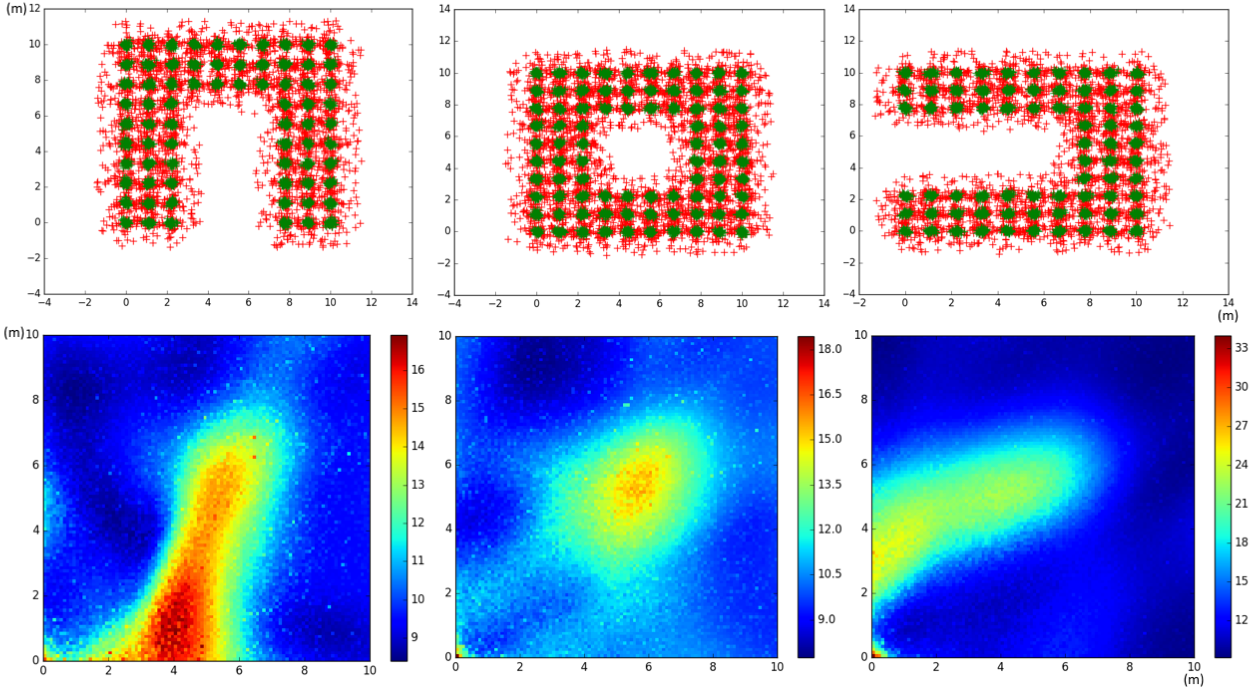}
 \caption{Simulated data collected from random pushes in Box2D and then lesioned in different ways (left). On the right are the corresponding heat maps depicting forward model model predictive uncertainty in different regions of the state space (right). Starting locations are shown as green crosses, positions after pushes are shown as red crosses.}
\label{fig:ensembleResults}
\end{figure}

To show that uncertainty rises when the amount of available data is limited we also gathered data from randomised pushes in Box-2D \cite{catto2011box2d}. Then we lesioned the data for various parts of the state space, and trained the E-MDN model. The uncertainty should be higher in the lesioned parts of space, and this is what we see in Figure~\ref{fig:ensembleResults}, in which the average uncertainty for a given state was obtained via marginalisation over the action space at the given state (average uncertainty for a given box state $\mathbf{x}$ was calculated using Monte-Carlo action samples).

\subsection{Experiment 2:}
\begin{figure}[t]
\centering
\includegraphics[width=0.98\columnwidth]{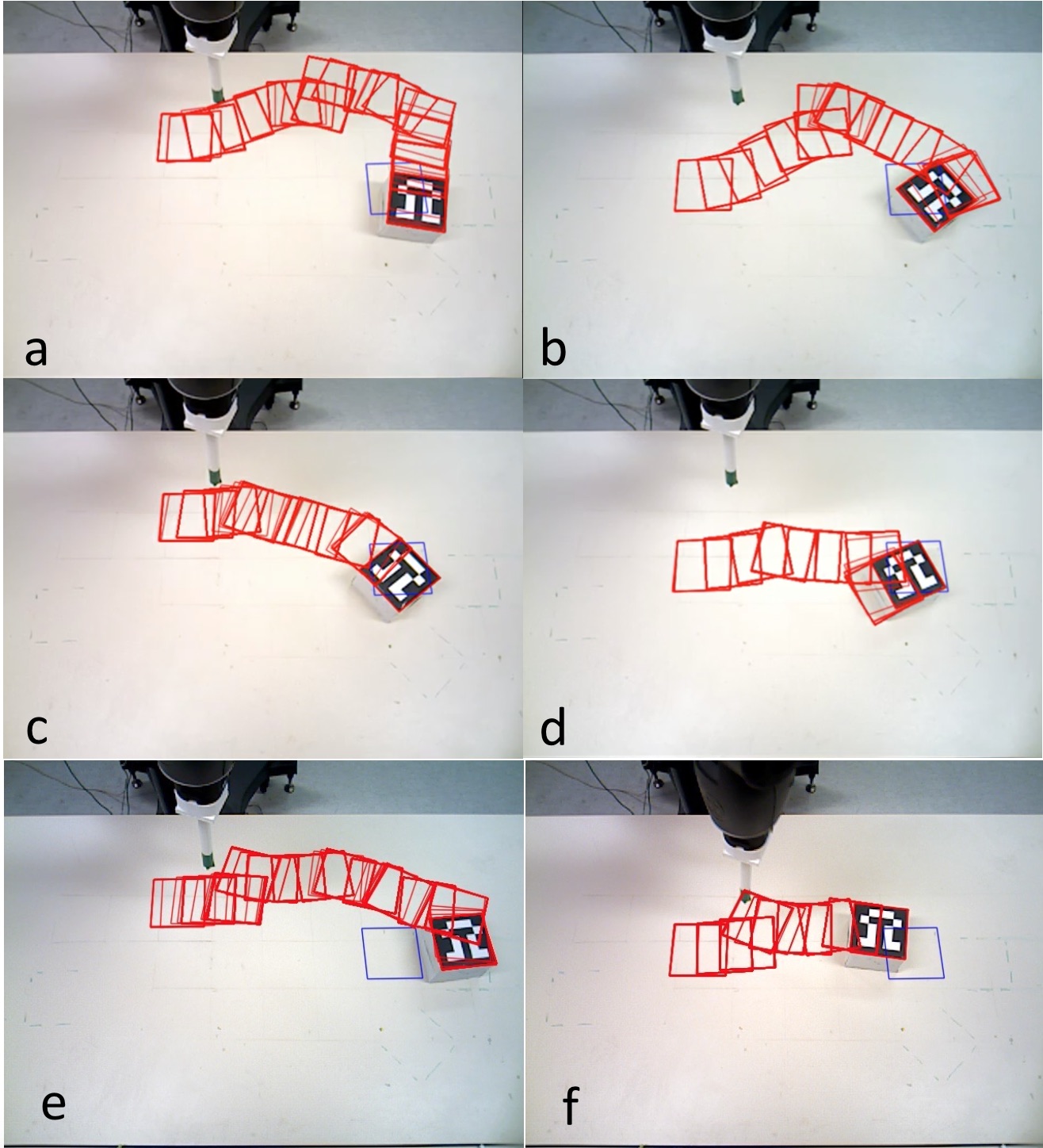}
\caption{\label{fig:push-runs}  Push sequences from two positions with MPPI. The first row contains results for Box2D (a and b), the second row contains the results for E-MDN (c and d) and the third row contains results for GP (e and f). Red frames show the starting and intermediate positions, the blue frame shows the goal. In figures a and c the goal has the same orientation as but different position from the initial box pose, whereas in b and d the goal is orientated 90 degrees anti-clockwise with respect to the initial box pose. In our experiments MPPI-E-MDN reached the goal with fewer pushes.}
\end{figure}

We performed experiments with the real robot, using the MPPI planner to plan with the learnt models from Experiment 1, and also with a physics simulator suitable for planar pushing called Box-2D \cite{catto2011box2d}. We use this to investigate whether the push planning framework can be combined with a variety of forward models to achieve many-step push manipulations. Some push sequences are visualised in Figure~\ref{fig:push-runs}. These show that both learning methods terminate with positions close to, but not at, the goal, in terms of both orientation and position. Table~\ref{tab:robot-data} shows that both substantially reduce the cost in paired trials, and that there is no clear difference in performance between the different predictors underpinning MPPI. The cost parameters for the MPPI were chosen to be $\mathbf{Q} = diag(1.5,1.5,0.01)$, $\gamma = 0$, $\mathbf{R} = \mathbf{0}$. The optimisation horizon was set to $T=2$ and the number of sampled trajectory rollouts was chosen to be $N=150$, $L=20$, with $h=1$, $\rho=1.0$, and $\Delta t=0.05$.

\begin{table}
\begin{tabular}{|c|c|c|c|c|}  \hline 
Treatment & Starting pose & Initial cost & Final cost  & Steps \\ \hline
MPPI-Box2D & Pose 1 & 0.800 & 0.112  & 13 \\ \hline
MPPI-E-MDN & Pose 1 & 0.795 & 0.057 & 8  \\ \hline
MPPI-GP & Pose 1 & 0.764 & 0.255 & 12  \\ \hline
MPPI-Box2D & Pose 2 & 0.766 & 0.097  & 12 \\ \hline
MPPI-E-MDN & Pose 2 & 0.768 & 0.079 & 8 \\ \hline
MPPI-GP & Pose 2 & 0.729 & 0.072 & 9  \\ \hline
\end{tabular}
\caption{\label{tab:robot-data} The cost is a weighted average of the position error (metres) and orientation error (radians).}
\end{table}

\subsection{Experiment 3:}
\begin{figure}[!th]
\centering
\includegraphics[width=0.4\textwidth]{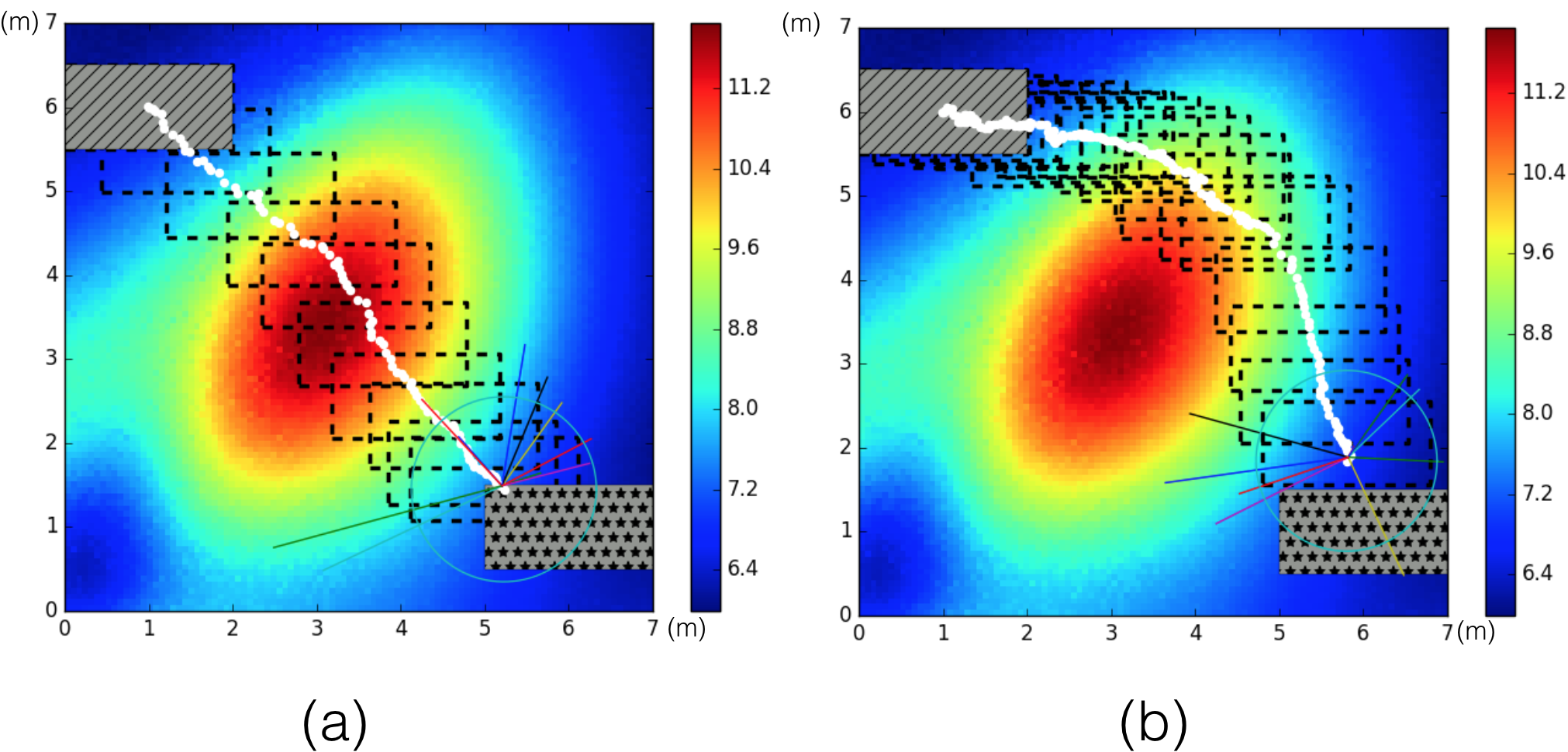}
    \caption{Uncertainty averse pushing (Algorithm 1). The top and bottom grey rectangles depict the start and goal locations, respectively. Dashed boxes in black show sub-sampled box locations along the trajectory in white. In simulation the box is pushed at its CoR, so it will not rotate. (a) Without an uncertainty penalty, the box follows a straight line towards the target. (b) Penalizing uncertainty, the box avoids it. The heuristic cost is defined such that uncertainty acts as a penalty for 150 pushes, and is set to zero afterwards ($\gamma = 0$). For this experiment $\gamma = 115$, $\mathbf{Q}  = diag(0.5, 0.5, 0.5)$, $h = 2.0$, $\rho = 2.0$, $T = 2$, $\Delta t = 0.05$, $N = 10, L = 0$. As before, the multi-coloured lines are forward sampled trajectories from the current box state, and the circle radius represents current uncertainty in dynamics.}
    \label{fig:unc_tradeoff_experiment_on}
\end{figure}

Utilising Algorithm \ref{alg:algorithm_1}, together with the cost function defined by Equation~\ref{eq:mppi_cost} set to penalise uncertainty for 150 pushes, and having the uncertainty penalty switched off there after. The aim of this experiment was to show in simulation that, with the right trade-off between the goal and the uncertainty gains, a box can be pushed to a desired location while avoiding regions with high uncertainty. The E-MDN was trained with 261 pushes collected from simulation. The parameters of the model were $M=10$, in which each MDN member of the ensemble had a single hidden layer with 25 units, and the number of mixtures was set to $K=1$. The MPPI parameters were set to $\mathbf{Q} = diag(0.5,0.5,0.5)$, $h = 2.0$, $\rho = 2.0$, $T = 2$, $\Delta t = 0.05$, $N = 10$, $L = 0$ (exploration decay not utilised) and the uncertainty penalty was set to $\gamma = 115$ for 150 pushes, and then $\gamma = 0$ afterwards. The results for this first experiment are shown by Figures~\ref{fig:unc_tradeoff_experiment_on} (a) and (b), in which two distinct trajectories are obtained as a result of either penalising or not penalising for model uncertainty.

\subsection{Experiment 4:}
\begin{SCfigure}[][!h]
\includegraphics[width=0.48\columnwidth]{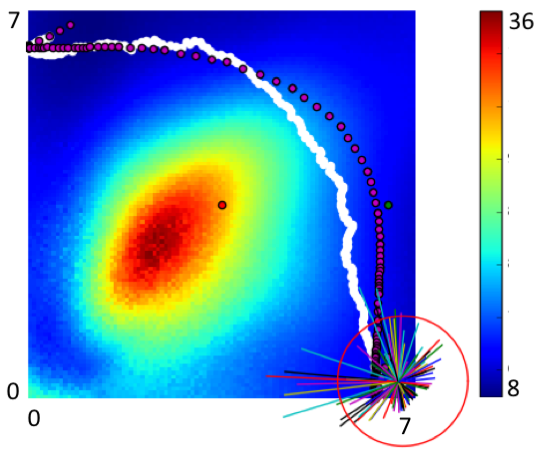}
  \caption{Uncertainty averse pushing (Algorithm 2). The red dots are the way-points to be followed and are generated by Algorithm 2. Algorithm 1 then attempts to follow these, producing the actual trajectory (white line). Forward sampled trajectories from the current box state are shown as multi-coloured lines, and the circle radius represents current region uncertainty.}
   \label{fig:traj_follow_algorithm2}
\end{SCfigure}

Finally, this experiment makes use of Algorithm \ref{alg:algorithm_2}, which performs optimisation to find a low uncertainty cost trajectory first. This low uncertainty cost trajectory is then followed by Algorithm \ref{alg:algorithm_1} push controller, but this time, we do not need to penalise uncertainty in its running cost, since the trajectory has already been optimised for that. The results for this experiment are shown in Fig \ref{fig:traj_follow_algorithm2}.

\section{DISCUSSION}
\label{discussion}
A push planning approach that uses a learnt forward model is presented. The push planner is also capable of taking into account the reliability of the learnt model. Initially we showed how a learnt forward model can be used by a real robot to push the object to a target location using the MPPI approach. Later, we showed the modification to this basic MPPI to accommodate predictive uncertainty in two ways. In the first algorithm the uncertainty is directly inserted into the MPPI cost function.  In the second formulation, a trajectory that is uncertainty averse is pre-computed using a path-integral update (Algorithm 2) and MPPI (Algorithm 1) is used to follow it.  We have shown that both algorithms exhibit the desired behaviour subject to tuning. In addition we have created the data gathering framework on a Baxter robot, and shown that E-MDNs and GPs produce very similar estimates of model uncertainty for real data. Experiments showed that Algorithm 1 works on the real robot, and that either one or both learning methods outpeformed a physics simulator, when used as the forward model for planning. \\

\bibliographystyle{plain}
\bibliography{bibliography/pushing,bibliography/opt_ctrl}

\end{document}